\documentclass{article}

\usepackage{arxiv}

\usepackage[utf8]{inputenc} 
\usepackage[T1]{fontenc}    
\usepackage{hyperref}       
\usepackage{url}            
\usepackage{booktabs}       
\usepackage{amsfonts}       
\usepackage{nicefrac}       
\usepackage{microtype}      
\usepackage{graphicx}
\usepackage{natbib}
\usepackage{doi}
\usepackage{tabularx}

\title{Violet Teaming in Practice: Lessons from a Pilot Expert Workshop on Governing Complex AI Systems}

\author{Steph Batalis\textsuperscript{1}, Kait Connors\textsuperscript{2}, Ben
C. Snyder\textsuperscript{3}, Vikram Venkatram\textsuperscript{1}, Josh
Wentzel\textsuperscript{3} \\
\\
\textsuperscript{1}Center for Security and Emerging Technology, Georgetown University, Washington D.C., United States \\
\textsuperscript{2} Independent Researcher, Washington D.C., United States \\
\textsuperscript{3} Biosecurity and Pandemic Policy Center, Bush School of Government \& Public Service, \\
Texas A\&M University, Washington D.C., United States
}

\renewcommand{\headeright}{Workshop Report}
\renewcommand{\undertitle}{Workshop Report}
\renewcommand{\shorttitle}{Violet Teaming in Practice}

\hypersetup{
pdftitle={Violet Teaming in Practice: Lessons from a Pilot Expert Workshop on Governing Complex AI Systems},
pdfsubject={cs.AI},
pdfauthor={Steph~Batalis, Ben C.~Snyder, Kait Connors, Vikram Venkatram, Josh Wentzel},
pdfkeywords={},
}

\begin{document}
\maketitle

\begin{abstract}
Violet teaming has been proposed as a method to integrate technical,
policy, and societal considerations into the evaluation of emerging
technologies. Here, we document a pilot expert workshop used to explore violet teaming as a governance methodology for biodefense-relevant AI capabilities and describe observations from the exercise. The workshop used a hypothetical technology scenario across three sequential stages---capability development, misuse potential, and
implementation---while incorporating technical, security, operational,
and policy perspectives. The sequential format provided an integrated
way to examine potential benefits, misuse risks, and real-world
implementation challenges, rather than considering these issues in
isolation. The exercise surfaced non-obvious misuse pathways, expanded
potential interventions beyond purely technical safeguards, and
highlighted operational and policy constraints that could affect
implementation. This report summarizes the workshop design, outputs, and lessons learned for applying violet teaming to the evaluation of
complex, dual-use technologies.
\end{abstract}

\section{Introduction}
Across domains, the rise of artificial intelligence (AI) has been
accompanied by fundamental questions about how to balance societal
benefits against potential negative impacts \citep{kapoor_societal_2024}. Developing and responsibly deploying AI systems is a complex task that requires keeping pace with rapid advances. AI governance strategies must coordinate between multiple concurrent priorities, including leveraging AI to solve pressing problems, understanding its unintended consequences, developing mitigation techniques to enhance safety, and
integrating systems into existing infrastructure \citep{bengio_managing_2024, nist_artificial_2023}. These challenges are particularly salient for emerging classes of AI systems, including agentic and semi-autonomous systems, where behavior emerges from interactions across technical and social environments \citep{chhabra_agentic_2026, chan_harms_2023}.

A number of methodologies exist to test AI systems and inform such governance. These approaches evaluate features such as system
performance, the efficacy of safety mitigations, or real-world implementation challenges \citep{ji_ai_2025}. Each of these methodologies, including red and blue teaming, capability evaluations, uplift studies, field studies, sandbox testing, and pilot programs, systematically evaluate a component of the AI development lifecycle \citep{frontier_model_forum_issue_2024, amironesei_assessing_2025}. However, because these approaches are typically applied to individual stages or dimensions of a system, they do not systematically capture how technical, policy, and implementation considerations interact across the full lifecycle of a technology.

A hybrid approach is needed to improve governance of complex sociotechnical systems, recognizing that AI tools cannot be evaluated
for their technical capabilities alone but must also be understood in context of their interaction with human behavior, organizational
practices, and institutional structures. First, some capabilities are needed regardless of their potential vulnerabilities. For these
high-demand applications, governance needs to move past simply identifying risks and benefits and towards charting solutions forward.
Second, challenges and limitations can surface between developers, security experts, implementers, users, and decision-makers if these perspectives are not integrated before deployment \citep{oduro_ai_2024}. Finally, even the most useful and low-risk capabilities do not
translate into real-world benefits if they cannot be implemented due to misalignments between actual and assumed needs, policy barriers, or resource limitations. As a result, the ability to systematically integrate perspectives across the technology lifecycle, and across technical and governance domains, is critical for identifying viable and
responsible deployment pathways \citep{chen_explainer_2024}.

Recent work has proposed violet teaming as a way to integrate these approaches \citep{titus_promise_2023}. It combines and builds on elements from two traditional methods: red-teaming, or adversarial vulnerability probing, and blue teaming, or problem-solving to develop mitigations to those problems \citep{zhang_red_2020, abuadbba_promise_2025}. Violet teaming extends these principles by evaluating not only the technology
itself but also the ways that people and societies work ``with, through, and because of'' AI systems \citep{titus_promise_2023}. The method's sociotechnical approach accounts for the broader policy, social, and governance decisions that are made in the process of integrating a technology into real-world systems and infrastructures. It acknowledges that making a technology safe requires integrating safety throughout its lifecycle, including by considering potential pitfalls during its development, in its intended use case(s), and in the wide variety of potential unintended use cases. Further, where red, blue, and purple teaming methods often include hands-on testing of systems, violet teaming can be a more theoretical exercise, providing a structured approach to analyze abstract or potential capabilities of future technologies and their implications. For these reasons, violet teaming could offer policymakers a more balanced, comprehensive view of governance and development pathways for emerging technologies.

This paper presents our reflections on the violet teaming concept, based on a workshop organized around violet teaming principles. While prior work has proposed violet teaming conceptually, to our knowledge it had
not yet been operationalized in practice. We decided to apply the approach by incorporating violet teaming principles into an expert workshop focused on AI capabilities for pathogen-agnostic biosurveillance, a strategy that involves continuously monitoring environmental and population samples for the presence of bacteria, viruses, fungi, or other biological entities \citep{gauthier_agnostic_2023}. By detecting and characterizing all potentially disease-causing pathogens,
rather than testing for a predefined list of known ones,
pathogen-agnostic biosurveillance represents a use case with the potential to enhance U.S. public health and biodefense efforts but which has also raised concerns about potential misuse \citep{kuiken_artificial_2023}. Beyond the domain-specific findings it generated, the workshop also provided an opportunity to observe how violet teaming principles shaped the exercise itself. This report therefore focuses on the workshop design, outputs,
and observations about how the violet teaming format supported integrated, decision-relevant consideration of capability development, misuse risks, and implementation challenges.

\section{Method: Incorporating Violet Teaming Elements into Workshop Design}

We incorporated the sociotechnical elements of violet teaming into our expert workshop by examining a hypothetical technology through multiple perspectives across its development-to-deployment lifecycle.

We designed the workshop to incorporate three elements of violet teaming: identifying beneficial capabilities, examining potential misuse, and developing mitigation and implementation strategies. To do so, we structured the exercise as three sequential sessions, summarized in Table 1. Each session used a narrative scenario that assigned a defined role, objective, resources, and constraints. In the first session, groups took the perspective of a hypothetical advisory committee tasked with identifying AI capabilities that would be most beneficial to Department of Defense biosurveillance programs. The AI capabilities discussed were hypothetical but grounded in current technical constraints and real-world needs. In the second session, groups adopted an adversarial perspective to examine how the same capabilities could be misused or exploited. In the final session, groups took on the perspective of government program managers tasked with considering how the proposed AI capabilities could be implemented while addressing the previously-identified vulnerabilities alongside resource, infrastructure, and resource needs.

\begin{figure}[htb]
    \centering
\includegraphics[width=1\linewidth]{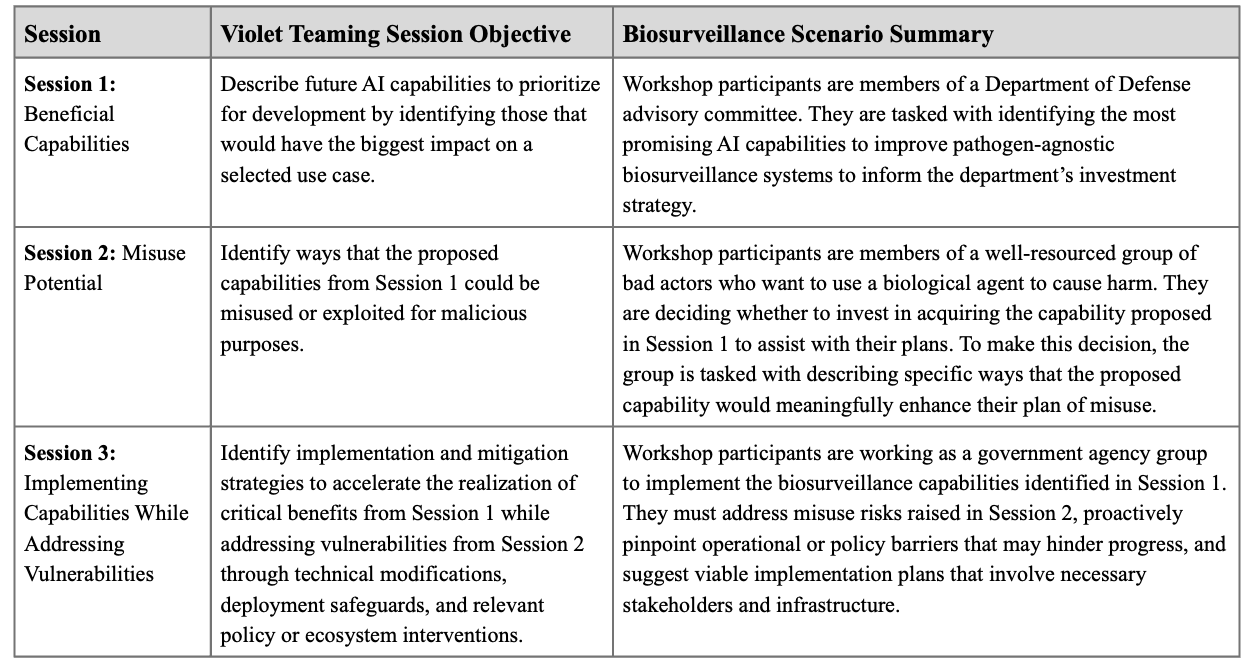}
    \caption{\textbf{Session objectives and scenario summaries for the sequential three-part violet teaming workshop.} Session 1 focused on identifying beneficial future AI capabilities, Session 2 examined potential misuse pathways, and Session 3 addressed implementation challenges and mitigation strategies. Participants adopted different perspectives across sessions to explore AI-enabled biosurveillance systems.}
\end{figure}

We chose to design the workshop so that each session builds on the previous one by following the same hypothetical AI capability throughout the workshop. In other words, the technology that groups proposed in the first session became the basis for the subsequent sessions focused on misuse and implementation. We hoped that this approach would provide a more holistic view of a technology, incorporating its technical features, intended and unintended use cases, and logistical and operational bottlenecks. We also hoped that this format would better reflect real-world governance challenges, which must move past simply identifying risks and benefits and instead address paths forward. Finally, following the same capability through the technology lifecycle gave participants a chance to evaluate how decisions made in an earlier phase of development impact later ones.

A defining characteristic of the violet teaming paradigm is its emphasis on including both technological and operational perspectives into emerging technology governance. We incorporated this approach into our workshop by including experts with practical experience across the development-to-implementation pipeline. Our goal was to bring together experts with complementary but distinct perspectives to ensure that discussions accounted for both technical feasibility and real-world constraints. We anticipated that combining the categories of expertise described in Table 2 would provide a more comprehensive basis for considering technical feasibility, misuse potential, and implementation constraints.
\begin{figure}[htb]
    \centering
\includegraphics[width=1\linewidth]{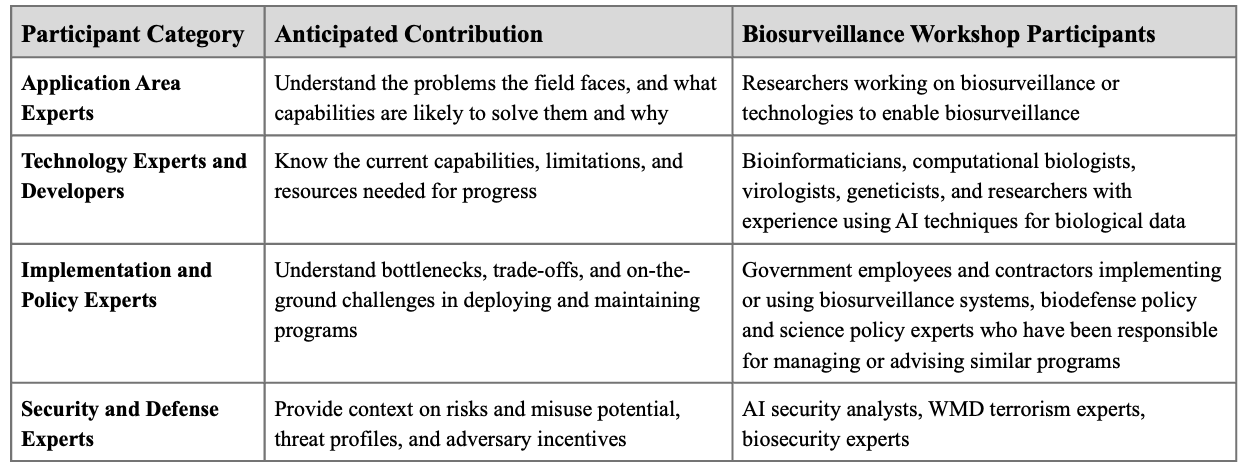}
    \caption{\textbf{Expert contributions to the violet teaming exercise.} This table describes categories of experts that may be particularly valuable to include in a violet teaming exercise (left column), how we anticipate they would contribute to the exercise (middle column) and how we included these categories in our biosurveillance-oriented pilot workshop (right column).}
\end{figure}

On the day of the workshop, we assigned experts to smaller groups of four to five people each, doing our best to ensure that each group contained at least one expert from each category in Table 2. Each session began by reviewing the provided scenario, followed by approximately one hour for groups to discuss the scenario and develop the specified session output. Moderators from the workshop organizing committee were assigned to each group to lead the scenario, answer questions or clarify details about the narrative, and document key discussion points and exercise outputs. Each session concluded with a brief facilitator-led debrief to capture observations about the exercise, clarify outstanding questions, and identify issues with the scenario or workshop structure. These debriefs also provided an opportunity to identify practical considerations for refining the workshop structure in future exercises.

The workshop design described here is only one of the many ways that violet teaming could be operationalized in practice. Our objective was to surface a wide range of ideas regarding AI capabilities, constraints, and governance considerations for biosurveillance, rather than to quantitatively assess the performance or risk of a given biosurveillance technology. We therefore prioritized exploratory scenarios that allowed for breadth and creativity in the exercise outputs.

This informed the decisions we made to prioritize exploratory, speculative scenarios that allowed for breadth and creativity in the responses. We do not intend for the methodology described here to be a prescriptive template for future iterations of violet teaming. Instead, we offer our observations from our violet teaming workshop to highlight key design considerations, trade-offs, and lessons learned that may inform future scenario-based evaluations.

\section{Workshop Observations: Applying Violet Teaming to Emerging Technology Governance}
Incorporating violet teaming principles provided a structured way to consider the same emerging technology from multiple technical, security, and operational perspectives. The violet teaming format helped connect these perspectives within a single exercise instead of treating them as separate questions. Rather than producing isolated insights, the format supported a more integrated view of the capabilities, risks, and implementation considerations associated with the technology under discussion.
The domain-specific technical recommendations generated during the workshop are not the focus of this report. Instead, this report documents our observations about how incorporating violet teaming principles shaped the workshop design and the resulting exercise outputs.
That being said, the workshop did generate high-quality, domain-specific outputs, including:
\begin{itemize}
    \item Identifying and prioritizing the concrete AI capabilities that would best address current limitations in biosurveillance pipelines, based on an assessment of real-world bottlenecks;
    \item Assessing risk potential while differentiating between less plausible and more realistic misuse scenarios, an important contribution to inform where research and mitigations should be directed to have the greatest impact; and
    \item Surfacing real-world implementation barriers, including policy, resource, and coordination challenges that would need to be addressed to support effective deployment.
\end{itemize}

\subsection{Violet teaming can generate integrated insights across the technology lifecycle}

Following the same hypothetical capability across sessions created a clear throughline from development to potential misuse and implementation. This allowed the exercise to surface a more comprehensive view of bottlenecks and opportunities and trace how the same technology might evolve through its lifecycle. The proposed recommendations changed at each phase as groups considered the problem from a different perspective, first focusing on prioritizing biodefense-relevant priorities, then how those capabilities may be misused, and finally on how those risks could be mitigated in practice, while accounting for anticipated process and policy requirements in implementing these mitigations.

This structure also shifted the focus from identifying risks in isolation to developing holistic, actionable recommendations. The workshop not only surfaced vulnerabilities, but also explored corresponding technical safeguards, policy interventions, and implementation strategies tied directly to the same capability. For example, the beneficial capabilities proposed in the first session were then stress-tested in subsequent sessions. In several instances, outputs from later sessions prompted revisions or reprioritization of ideas developed earlier in the exercise. If sessions had been held in isolation, there would not have been an opportunity to reconsider the capability as new technical, security, and implementation constraints were introduced.

This continuity distinguishes the format from exercises that examine benefits, risks, and implementation separately. By carrying the same hypothetical capability through three sequential sessions, the workshop supported an end-to-end assessment rather than producing separate lists of opportunities, threats, and potential solutions that are difficult to integrate.

\subsection{The workshop surfaced novel, non-obvious insights}
The violet teaming format also supported focused analysis of a specific technology rather than discussion of an entire technology category in the abstract. In our workshop, this focus helped surface risks and constraints specific to the capability under discussion rather than relying solely on commonly assumed risks associated broadly with AI-enabled biological technologies.

At the time of the workshop, public and policy discussions frequently cited concerns that AI-enabled biological tools could be misused to aid in creating biological weapons by designing more targeted, infectious, or otherwise dangerous pathogens \citep{national_academies_of_sciences_engineering_and_medicine_age_2025, carter_convergence_2023}. Yet, within the constraints of the defined technology and scenario, groups found that the AI capabilities under discussion were not particularly useful for designing harmful biological agents. In fact, taking on the perspective of a hypothetical malicious actor revealed that trying to incorporate the specified AI capabilities into a plan of misuse actually made those workflows less efficient and more complex than simpler alternatives. Instead, the exercise surfaced alternative misuse pathways that appeared more relevant to the specific biosurveillance capabilities under discussion, including potential data poisoning or cyber exploitation that could create a “smokescreen” of false positives to distract from a real attack or undermine confidence in the system.

The workshop demonstrated that considering a single technology’s potential application, risks, and implementation generated insights that challenged common assumptions about how emerging technologies are likely to be used or misused. This is an important step in developing and prioritizing solutions that are correctly targeted to the underlying problem. Violet teaming’s sociotechnical approach may be particularly valuable in correcting misaligned intuitions about where risks are most likely to materialize, especially in a rapidly evolving technological domain.

\subsection{The sequential format supported candid risk identification and constructive problem-solving}

An important feature of the sequential format was that it gave groups an explicit opportunity to revisit problems identified earlier in the exercise. We found that previewing the full sequence at the outset helped groups engage more fully with the misuse exercise. In particular, knowing in advance that there would be an opportunity to address identified vulnerabilities created space for groups to discuss those problems candidly during an earlier stage of the exercise. In this framing, potential risks are treated as inputs to a subsequent stage of the exercise rather than endpoint conclusions about the value or feasibility of the proposed capability. The misuse exercise therefore becomes a step toward making a potentially valuable capability safer and more feasible, rather than a standalone exercise in finding reasons it could go wrong.

The sequence also helped maintain a focus on solutions once vulnerabilities had been identified. Because the workshop began by asking groups to identify AI capabilities that could provide significant value for biosurveillance, it established a grounded, real-world reason to find ways to overcome limitations and preserve the technology’s potential utility. In subsequent sessions when the misuse exercise surfaced potential vulnerabilities, groups were incentivized to work on mitigating those vulnerabilities rather than treating the risks as a reason to dismiss the capability altogether.

This dynamic was particularly useful because formats that focus only on potential risks can lead to promising capabilities being dismissed prematurely, while formats that focus only on potential benefits can understate important vulnerabilities. If an exercise is perceived as likely to over-index on either benefits or risks, it may be more difficult to surface limitations with sufficient detail or nuance. The sequential structure instead created a deliberate path from identifying high-value capabilities, to examining how they could be misused, to considering how those vulnerabilities could be addressed through technical, operational, and policy interventions.

\subsection{The workshop surfaced real-world bottlenecks early}
We observed that organizing the workshop to incorporate violet teaming’s emphasis on sociotechnical factors helped to surface political, logistical, and bureaucratic barriers that technical experts and developers alone might overlook without the input of policy experts. 

In our workshop, including policy and implementation expertise in our exercise helped to surface context about navigating government processes and designing and managing federal programs. This included raising awareness of how practical barriers like data-sharing limitations, classification constraints, and funding instability shape the feasibility of deployment. For example, in one case policy experts explained that one theoretically promising proposal would require authorities that the government entity specified in the scenario did not possess. The proposal was therefore revised within the exercise to reflect a more feasible implementation pathway. 

This is especially valuable because recognizing real-world constraints early can help to guide future development away from implementation “dead ends” and towards more operationally feasible options. Building effective tools requires not only technical innovation, but policy support and early clarity on what actions can be realistically accomplished. Violet teaming’s integrated approach could help uncover misalignments between stakeholders and prioritize options that are both scientifically and operationally realistic. 

\subsection{The violet teaming format expanded the possible solution space considered in the exercise}
Violet teaming’s inclusion of both technical and non-technical considerations creates a larger solution space to explore when addressing identified vulnerabilities. We observed that considering multiple perspectives and intervention points through the violet teaming format enabled more comprehensive and realistic approaches to risk mitigation across technical, operational, and policy domains.

In our workshop, this resulted in layered approaches to risk mitigation that combined technical safeguards (e.g., access controls, anomaly detection, and cybersecurity measures) and system-level interventions (e.g., data sharing protocols, program management mechanisms, and interagency coordination strategies). In one example, a group identified a theoretical misuse pathway that involves gaining unauthorized access to a biosurveillance system and manipulating it to display a false laboratory test result. While technical safeguards can reduce the likelihood of such attacks, cybersecurity practices alone cannot entirely eliminate such vulnerabilities or guarantee a fully secure system. Instead, the group proposed a verification protocol that involves picking up the phone and calling the laboratory to directly confirm high-consequence results. This low-tech, procedural suggestion only occurred when participants looked past the AI system itself and also considered how it is integrated into the desired deployment context.

This broader framing matters because many AI risk analyses are narrowly focused on technical failure modes, with solutions framed primarily in terms of modifications to the underlying system or model. This can make certain vulnerabilities appear difficult or impossible to mitigate when purely technical solutions cannot fully address the underlying problem. The violet teaming format instead creates space to consider safeguards at the technical, operational, and policy levels. 

\section{Design Considerations for Future Violet Teaming Exercises}
In addition to the observations described above, the workshop also highlighted how specific design choices shaped the exercise and its outputs. Here, we share our thoughts on key decisions that shaped the workshop’s outcomes. These broader design considerations are not intended to constitute a systematic evaluation of scenario-based methods, but are rather offered as practical considerations for others adapting violet teaming or similar scenario-based approaches to different contexts.

\subsection{Narrative, role-based scenarios anchor abstract discussions in concrete decision contexts}

Framing each session as a story-based scenario with defined roles, constraints, and objectives, rather than a general discussion of a given technology, helped align workshop participants around a shared understanding of their task and enhanced overall engagement. Providing detailed assumptions and constraints for each hypothetical scenario helped minimize unproductive “what if” debates and set boundaries for the scope of the discussion. This format is not a requirement for violet teaming, but we found that it encouraged more specific, actionable outputs by situating technical and policy questions within realistic decision-making contexts.

\subsection{Scenario specificity can shape the breadth and focus of exercise outputs}
Violet team designers can modify the level of detail and world-building in the provided narrative to produce outcomes that align with their goals for the exercise. If the exercise’s objective is to generate a wide range of ideas and surface many avenues for further exploration, open-ended scenarios may provide more creative freedom and allow participants to draw on their own areas of expertise. If the goal is to stress-test a particular technology or implementation pathway, then constrained narratives may offer a concrete starting point, anchor groups to the same reference point, and enable clearer comparisons across groups. We observed this dynamic during the second session, when our decision to leave certain details unspecified sometimes caused groups to spend time debating what to focus on. At the same time, the most valuable and novel insights occurred during these open-ended discussions when participants explored a wider range of possibilities than we had imagined during session design. 

\subsection{Interdisciplinary groups benefit from shared baseline knowledge}
The interdisciplinary nature of the exercise meant that all participants sometimes would be engaging with subjects outside their area of expertise. Without a shared baseline of technical, policy, and operational context, conversations occasionally wandered or stayed at a higher level of abstraction. This suggests that spending additional time establishing key concepts, for example through an introductory session or more thorough background materials, would improve engagement and enhance the flow of ideas across disciplines.

\subsection{Facilitation and structured debriefs help document exercise outputs}
Facilitation and structured, moderator-guided debriefs provided opportunities to clarify scenario assumptions, capture observations that arose during the exercise, and identify issues for consideration in future iterations. These mechanisms were particularly useful for documenting questions or ideas that fell outside the immediate scope of a session. For future exercises, facilitators may wish to incorporate dedicated mechanisms for capturing such observations alongside the primary exercise outputs.

\section{Additional Opportunities for Violet Teaming}
Taken together, our workshop observations suggest that violet teaming may be a useful strategy to develop a fuller picture of a technology’s sociotechnical implications and promote responsible deployment. While our workshop focused on AI-enabled biosurveillance, the benefits of the approach are not specific to this application domain. Instead, they reflect properties of the violet teaming method itself. More broadly, structured, role-based scenarios may help bridge disciplinary divides by anchoring discussions in concrete, shared contexts. While violet teaming is not a replacement for other methods of assessing AI safety, including traditional red and blue teaming, performance on safety benchmarks, and human uplift studies, it is a promising addition that widens the aperture of identifiable vulnerabilities and solutions.

Based on our pilot workshop, we observed two characteristics that may make a technology domain particularly well-suited to the violet teaming approach. Firstly, we suggest prioritizing technologies for which there is a clear demand for the capability regardless of potential risks, either because the benefits outweigh the risks or because deployment is likely to occur regardless of any risk assessments. For these high-value applications, violet teaming’s solutions-oriented approach can help to identify viable paths forward. Secondly, violet teaming may be especially useful when a technology is at an intermediate stage of maturity where design choices and policy frameworks can still be influenced. Violet teaming offers an opportunity to consider the design of technical and social systems before a technology is widely deployed. However, it functions best when the technology is sufficiently advanced to ground assessments in reality. Our workshop observations suggest that applications with a similar profile may benefit most from future violet teaming exercises.

Violet teaming is inherently flexible, allowing it to be tailored to suit a variety of contexts, goals, and participants. Technology developers could violet team their own prototype technologies to inform design and deployment decisions. Government agencies could violet team a particular development or acquisition program. And regulators could perform violet teaming to better align regulations with anticipated risks, surfacing both unexpectedly burdensome requirements and possible gaps. These iterations of the method could be applied across domains, from assessments of agentic AI capabilities, to integrating decision-support tools into clinical settings, to forecasting the risks of frontier AI systems.

\subsection{Discussion}
Our operationalization of violet teaming principles in a scenario-based, expert workshop illustrates the potential benefits of the approach as a method to explore complex, emerging technologies through a sociotechnical lens. In particular, our observations suggest that the method can help address persistent gaps in technology governance, including the fragmentation between technical analysis and policy design. 

Beyond violet teaming itself, this pilot workshop highlights broader implications related to how key design choices shaped the workshop’s outcomes. In particular, the use of sequential analysis, narrative-based scenarios, and interdisciplinary discussion appears to support novel and comprehensive insights. These design features may be applicable to other domains where emerging technologies introduce complex, cross-cutting risks and opportunities.

As technology continues to advance to solve high-stakes problems, new developments are complicated by the possibility that solutions could be misused or circumvented. Violet teaming offers a promising model to address other dual-use or high-stakes technologies. It may be especially relevant for the governance of agentic AI systems, where system behavior depends on complex interactions between models, users, and operational environments. Future work should test and refine the approach in additional domains, and adapt it based on context-specific needs. Exercise outputs are likely to vary depending on topic selection, expert composition, and scenario design, and additional evidence is needed to assess how violet teaming performs in a range of environments.

Violet teaming can help policymakers and developers alike anticipate challenges, implement security-by-design, and identify pathways to responsible implementation. We hope this work provides a foundation for further iterations to support responsible technology development and governance. As future efforts build on this initial pilot, we look forward to seeing how the broader community refines the method, expands its use cases, and strengthens its value as a decision-support tool.

\section{Acknowledgements}
The authors would like to thank Brandon Behlendorf and Alexander Titus for providing input on the workshop design, Gerry Parker for support during the grant application process, and workshop participants for their valuable insights.

\section{Funding Statement}
This material is based upon work supported by the National Science Foundation under Grant No. 2427760 as part of the Responsible Design, Development, and Deployment of Technologies (ReDDDoT) program. 

\section{Author Contributions}
JW, SB, BS, VV, and KC jointly contributed to project conceptualization, design, interpretation, and manuscript writing. JW secured financial support for the project.

\bibliographystyle{abbrvnat}
\bibliography{references}  

@techreport{carter_convergence_2023,
	title = {The {Convergence} of {Artificial} {Intelligence} and the {Life} {Sciences}},
	url = {https://www.nti.org/analysis/articles/the-convergence-of-artificial-intelligence-and-the-life-sciences/},
	language = {en},
	urldate = {2024-08-19},
	institution = {Nuclear Threat Initiative},
	author = {Carter, Sarah R. and Wheeler, Nicole E. and Chwalek, Sabrina and Isaac, Christopher R. and Yassif, Jaime},
	month = oct,
	year = {2023},
}

@techreport{kuiken_artificial_2023,
	title = {Artificial {Intelligence} in the {Biological} {Sciences}: {Uses}, {Safety}, {Security}, and {Oversight}},
	url = {https://crsreports.congress.gov/product/pdf/R/R47849},
	number = {R47849},
	institution = {Congressional Research Service},
	author = {Kuiken, Todd},
	month = nov,
	year = {2023},
}

@misc{ji_ai_2025,
	title = {{AI} {Safety} {Evaluations}: {An} {Explainer}},
	shorttitle = {{AI} {Safety} {Evaluations}},
	url = {https://cset.georgetown.edu/article/ai-safety-evaluations-an-explainer/},
	language = {en-US},
	urldate = {2026-04-09},
	journal = {Center for Security and Emerging Technology},
	author = {Ji, Jessica and Venkatram, Vikram and Batalis, Steph},
	month = may,
	year = {2025},
}

@article{chhabra_agentic_2026,
	title = {Agentic {AI} {Security}: {Threats}, {Defenses}, {Evaluation}, and {Open} {Challenges}},
	volume = {14},
	copyright = {https://creativecommons.org/licenses/by/4.0/legalcode},
	issn = {2169-3536},
	shorttitle = {Agentic {AI} {Security}},
	url = {https://ieeexplore.ieee.org/document/11447227/},
	doi = {10.1109/ACCESS.2026.3675554},
	urldate = {2026-04-09},
	journal = {IEEE Access},
	author = {Chhabra, Anshuman and Datta, Shrestha and Nahin, Shahriar Kabir and Mohapatra, Prasant},
	year = {2026},
	pages = {49455--49482},
}

@book{national_academies_of_sciences_engineering_and_medicine_age_2025,
	address = {Washington (DC)},
	title = {The {Age} of {AI} in the {Life} {Sciences}: {Benefits} and {Biosecurity} {Considerations}},
	copyright = {Copyright 2025 by the National Academy of Sciences. All rights reserved.},
	isbn = {978-0-309-73335-9},
	shorttitle = {The {Age} of {AI} in the {Life} {Sciences}},
	url = {http://www.ncbi.nlm.nih.gov/books/NBK614593/},
	language = {eng},
	urldate = {2026-04-09},
	publisher = {National Academies Press (US)},
	author = {{National Academies of Sciences, Engineering, and Medicine}},
	year = {2025},
	pmid = {40392968},
}

@misc{kapoor_societal_2024,
	title = {On the {Societal} {Impact} of {Open} {Foundation} {Models}},
	url = {http://arxiv.org/abs/2403.07918},
	doi = {10.48550/arXiv.2403.07918},
	urldate = {2026-04-09},
	publisher = {arXiv},
	author = {Kapoor, Sayash and Bommasani, Rishi and Klyman, Kevin and Longpre, Shayne and Ramaswami, Ashwin and Cihon, Peter and Hopkins, Aspen and Bankston, Kevin and Biderman, Stella and Bogen, Miranda and Chowdhury, Rumman and Engler, Alex and Henderson, Peter and Jernite, Yacine and Lazar, Seth and Maffulli, Stefano and Nelson, Alondra and Pineau, Joelle and Skowron, Aviya and Song, Dawn and Storchan, Victor and Zhang, Daniel and Ho, Daniel E. and Liang, Percy and Narayanan, Arvind},
	month = feb,
	year = {2024},
	note = {arXiv:2403.07918 [cs]},
}

@article{zhang_red_2020,
	title = {Red {Teaming} the {Biological} {Sciences} for {Deliberate} {Threats}},
	volume = {32},
	issn = {0954-6553},
	url = {https://doi.org/10.1080/09546553.2018.1457527},
	doi = {10.1080/09546553.2018.1457527},
	number = {6},
	urldate = {2026-04-09},
	journal = {Terrorism and Political Violence},
	publisher = {Routledge},
	author = {Zhang, Lisa and Gronvall, Gigi Kwik},
	month = aug,
	year = {2020},
	note = {\_eprint: https://doi.org/10.1080/09546553.2018.1457527},
	pages = {1225--1244},
}

@misc{abuadbba_promise_2025,
	title = {From {Promise} to {Peril}: {Rethinking} {Cybersecurity} {Red} and {Blue} {Teaming} in the {Age} of {LLMs}},
	shorttitle = {From {Promise} to {Peril}},
	url = {http://arxiv.org/abs/2506.13434},
	doi = {10.48550/arXiv.2506.13434},
	urldate = {2026-04-09},
	publisher = {arXiv},
	author = {Abuadbba, Alsharif and Hicks, Chris and Moore, Kristen and Mavroudis, Vasilios and Hasircioglu, Burak and Goel, Diksha and Jennings, Piers},
	month = jun,
	year = {2025},
	note = {arXiv:2506.13434 [cs]
version: 1},
}

@techreport{nist_artificial_2023,
	address = {Gaithersburg, MD},
	title = {Artificial {Intelligence} {Risk} {Management} {Framework} ({AI} {RMF} 1.0)},
	url = {http://nvlpubs.nist.gov/nistpubs/ai/NIST.AI.100-1.pdf},
	doi = {10.6028/NIST.AI.100-1},
	language = {en},
	number = {NIST AI 100-1},
	urldate = {2026-04-10},
	institution = {National Institute of Standards and Technology (U.S.)},
	author = {{NIST}},
	month = jan,
	year = {2023},
	pages = {NIST AI 100--1},
}

@techreport{oduro_ai_2024,
	type = {Policy {Brief}},
	title = {{AI} {Governance} {Needs} {Sociotechnical} {Expertise}},
	url = {https://datasociety.net/library/ai-governance-needs-sociotechnical-expertise/},
	language = {en},
	institution = {Data \& Society},
	author = {Oduro, Serena and Kneese, Tamara},
	month = may,
	year = {2024},
}

@techreport{chen_explainer_2024,
	type = {Policy {Brief}},
	title = {Explainer: {A} {Sociotechnical} {Approach} to {AI} {Policy}},
	url = {https://datasociety.net/library/a-sociotechnical-approach-to-ai-policy/},
	language = {en},
	institution = {Data \& Society},
	author = {Chen, Brian J and Metcalf, Jacob},
	month = may,
	year = {2024},
}

@techreport{frontier_model_forum_issue_2024,
	title = {Issue {Brief}: {Preliminary} {Taxonomy} of {Pre}-{Deployment} {Frontier} {AI} {Safety} {Evaluations}},
	shorttitle = {Issue {Brief}},
	url = {https://www.frontiermodelforum.org/updates/issue-brief-preliminary-taxonomy-of-pre-deployment-frontier-ai-safety-evaluations/},
	language = {en-US},
	urldate = {2026-04-10},
	institution = {Frontier Model Forum},
	author = {{Frontier Model Forum}},
	month = dec,
	year = {2024},
}

@techreport{amironesei_assessing_2025,
	address = {Gaithersburg, MD},
	title = {Assessing risks and impacts of {AI} ({ARIA}) : pilot evaluation report},
	shorttitle = {Assessing risks and impacts of {AI} ({ARIA})},
	url = {https://nvlpubs.nist.gov/nistpubs/ai/NIST.AI.700-2.pdf},
	doi = {10.6028/NIST.AI.700-2},
	language = {en},
	number = {NIST AI 700-2},
	urldate = {2026-04-10},
	institution = {National Institute of Standards and Technology},
	author = {Amironesei, Razvan and Godil, Afzal and Greenberg, Craig and Greene, Kristen and Hall, Patrick and Jensen, Theodore and Fiscus, Jonathan and Schulman, Noah},
	month = nov,
	year = {2025},
}

@article{bengio_managing_2024,
	title = {Managing extreme {AI} risks amid rapid progress},
	volume = {384},
	url = {https://www.science.org/doi/abs/10.1126/science.adn0117},
	doi = {10.1126/science.adn0117},
	number = {6698},
	urldate = {2026-04-10},
	journal = {Science},
	publisher = {American Association for the Advancement of Science},
	author = {Bengio, Yoshua and Hinton, Geoffrey and Yao, Andrew and Song, Dawn and Abbeel, Pieter and Darrell, Trevor and Harari, Yuval Noah and Zhang, Ya-Qin and Xue, Lan and Shalev-Shwartz, Shai and Hadfield, Gillian and Clune, Jeff and Maharaj, Tegan and Hutter, Frank and Baydin, Atılım Güneş and McIlraith, Sheila and Gao, Qiqi and Acharya, Ashwin and Krueger, David and Dragan, Anca and Torr, Philip and Russell, Stuart and Kahneman, Daniel and Brauner, Jan and Mindermann, Sören},
	month = may,
	year = {2024},
	pages = {842--845},
}

@inproceedings{chan_harms_2023,
	title = {Harms from {Increasingly} {Agentic} {Algorithmic} {Systems}},
	url = {http://arxiv.org/abs/2302.10329},
	doi = {10.1145/3593013.3594033},
	urldate = {2026-04-10},
	booktitle = {2023 {ACM} {Conference} on {Fairness} {Accountability} and {Transparency}},
	author = {Chan, Alan and Salganik, Rebecca and Markelius, Alva and Pang, Chris and Rajkumar, Nitarshan and Krasheninnikov, Dmitrii and Langosco, Lauro and He, Zhonghao and Duan, Yawen and Carroll, Micah and Lin, Michelle and Mayhew, Alex and Collins, Katherine and Molamohammadi, Maryam and Burden, John and Zhao, Wanru and Rismani, Shalaleh and Voudouris, Konstantinos and Bhatt, Umang and Weller, Adrian and Krueger, David and Maharaj, Tegan},
	month = jun,
	year = {2023},
	note = {arXiv:2302.10329 [cs]},
	pages = {651--666},
}

@misc{titus_promise_2023,
	title = {The {Promise} and {Peril} of {Artificial} {Intelligence} - {Violet} {Teaming} {Offers} a {Balanced} {Path} {Forward}},
	url = {http://arxiv.org/abs/2308.14253},
	doi = {10.48550/arXiv.2308.14253},
	urldate = {2026-05-04},
	publisher = {arXiv},
	author = {Titus, Alexander J. and Russell, Adam H.},
	month = aug,
	year = {2023},
	note = {arXiv:2308.14253 [cs]},
}

@article{gauthier_agnostic_2023,
	title = {Agnostic {Sequencing} for {Detection} of {Viral} {Pathogens}},
	volume = {36},
	url = {https://journals.asm.org/doi/10.1128/cmr.00119-22},
	doi = {10.1128/cmr.00119-22},
	number = {1},
	urldate = {2026-10-26},
	journal = {Clinical Microbiology Reviews},
	publisher = {American Society for Microbiology},
	author = {Gauthier, Nick P. G. and Chorlton, Samuel D. and Krajden, Mel and Manges, Amee R.},
	month = feb,
	year = {2023},
	pages = {e00119--22},
}

\end{document}